\documentclass[]{bytedance_seed}

\usepackage[toc,page,header]{appendix}

\usepackage{minitoc}
\usepackage{booktabs}

\usepackage{tabularx}
\usepackage{tabularx}
\usepackage{adjustbox}

\title{ BABE: Biology Arena BEnchmark}

\author[1,2,*]{Junting Zhou}
\author[1, *]{Jin Chen}
\author[2, *]{Linfeng Hao}
\author[1*]{Denghui Cao}
\author[1]{Zheyu Wang}
\author[1]{Qiguang Chen}
\author[1]{Chaoyou Fu}
\author[1]{Jiaze Chen}
\author[1]{Yuchen Wu}
\author[1]{Ge Zhang}

\author[1, \dagger]{Mingxuan Wang}
\author[1, \dagger]{Wenhao Huang}
\author[2, \dagger]{Tong Yang}

\affiliation[1]{ByteDance Seed}
\affiliation[2]{Peking University}

\contribution[*]{Work done at ByteDance Seed}
\contribution[\dagger]{Corresponding authors}

\abstract{
The rapid evolution of large language models (LLMs) has expanded their capabilities from basic dialogue to advanced scientific reasoning. However, existing benchmarks in biology often fail to assess a critical skill required of researchers: the ability to integrate experimental results with contextual knowledge to derive meaningful conclusions. To address this gap, we introduce \textbf{BABE} (Biology Arena BEnchmark), a comprehensive benchmark designed to evaluate the experimental reasoning capabilities of biological AI systems. BABE is uniquely constructed from peer-reviewed research papers and real-world biological studies, ensuring that tasks reflect the complexity and interdisciplinary nature of actual scientific inquiry. BABE challenges models to perform causal reasoning and cross-scale inference. Our benchmark provides a robust framework for assessing how well AI systems can reason like practicing scientists, offering a more authentic measure of their potential to contribute to biological research.
}

\date{\today}
\correspondence{Junting Zhou at \email{juntingzhou@stu.pku.edu.cn}; Tong Yang at \email{yangtong@pku.edu.cn}}

\begin{document}
\maketitle


\section{Introduction}

The evolution of large language models (LLMs) has witnessed a paradigm shift from basic conversational capabilities to advanced reasoning functionalities. Early-generation models excelled at generating coherent chat-style responses, but modern foundation models have expanded into \textbf{scientific research capabilities}—including hypothesis generation, data analysis, and experimental design. This shift has drawn significant attention to evaluating LLMs’ performance in specialized scientific domains, particularly biology, where complex experimental data and interdisciplinary knowledge demand more than trivial pattern recognition.

A critical yet underdeveloped aspect of assessing biological AI systems is their ability to reason based on experimental results and contextual background, which is a core skill for biological researchers. For instance, interpreting a Western blot image to infer protein expression changes requires integrating visual data (for example: band intensity, loading controls) with experimental context (for example: treatment conditions, cell lines) and domain knowledges. This kind of problem is a challenge even for the strongest current SOTA models. 

However, existing benchmarks rarely test this integrated reasoning ability, instead focusing on isolated tasks like sequence classification or structure prediction.

\begin{figure*}[h!]
    \centering
    \includegraphics[width=0.90\linewidth]{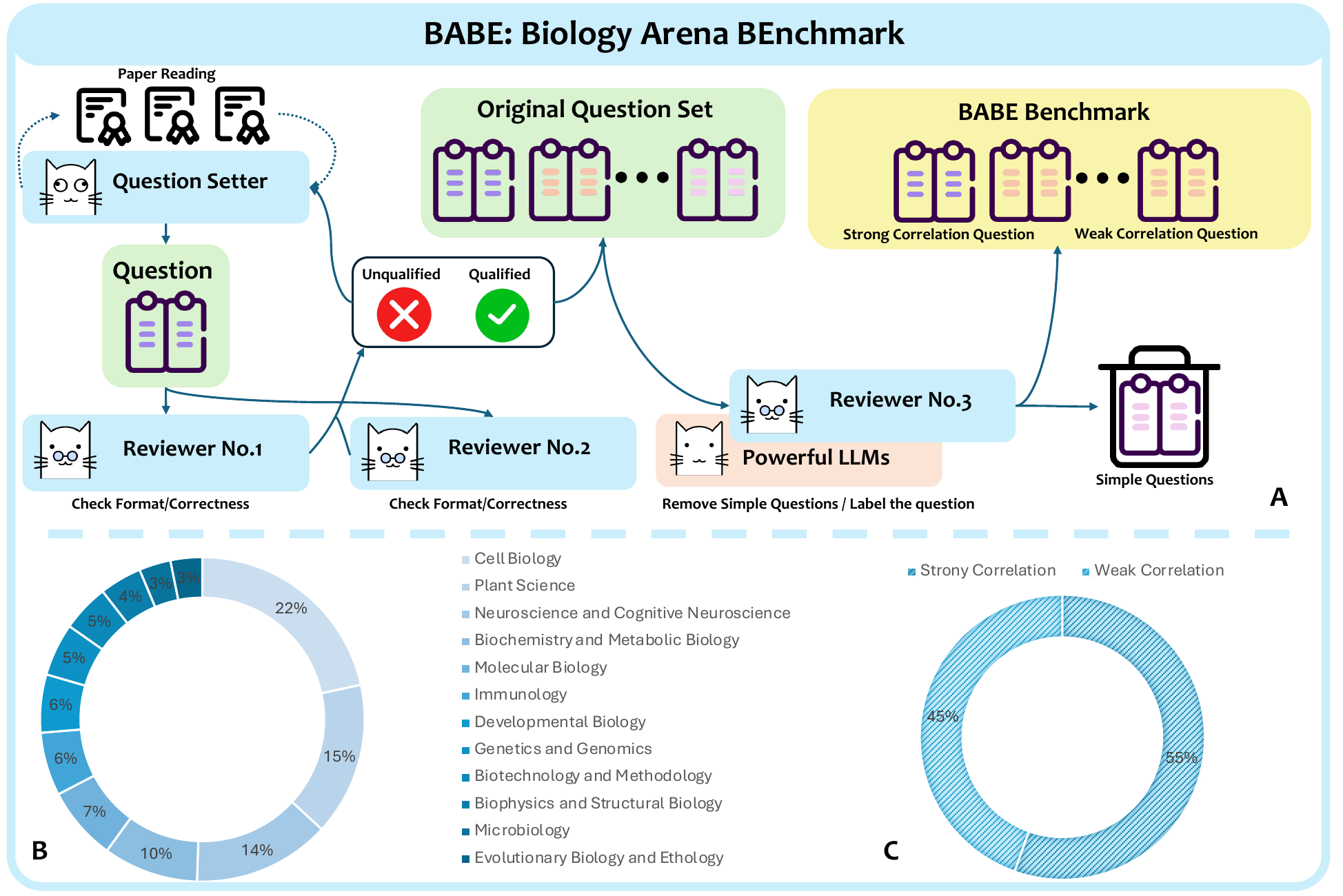}
    \caption{Overview of the BABE (Biology Arena BEnchmark) construction and composition. (A) The multi-stage annotation pipeline for constructing the BABE benchmark. (B) The disciplinary distribution of questions in the BABE benchmark, covering 12 subfields of biology. (C) The proportion of strong-correlation (45\%) and weak-correlation (55\%) questions in the final BABE benchmark.}
    \label{fig:main-fig}
\end{figure*}

To address these gaps, we develop BABE(Biology Arena BEnchmark\ref{fig:main-fig}), a benchmark specifically designed to evaluate biological AI systems’ experimental reasoning capabilities. Critically, all tasks in \textbf{BABE} are derived from peer-reviewed research papers and real-world biological studies. This ensures that the benchmark reflects the true complexity of biological research and tests models’ ability to reason like practicing scientists.

In summary, Our work makes three primary contributions:

\begin{itemize}
    \item Experimental Reasoning Focus: Unlike existing benchmarks, BABE centers on tasks that require models to integrate experimental results with contextual background to derivate biological conclusions. 
    \item High-Difficulty, Research-Derived Tasks: All tasks are adapted from peer-reviewed papers and designed to demand causal reasoning and cross-scale inference.
    \item Broad Domain Coverage: We cover major biological domains, using tasks from diverse subfield studies, enabling evaluation of model generalization across real-world biological research areas.
\end{itemize}

\section{Related Work}
\label{sec:related-work}

\subsection{Deep Research Agents}
The initial success of Large Language Models (LLMs) in natural language tasks was often hampered by inherent limitations in multi-step reasoning, knowledge cut-offs, and the propensity for hallucination \cite{hallucination}. This motivated a critical shift from static generation toward the development of sophisticated Deep Research Agents. These agents are designed to tackle complex, multi-turn informational research tasks by equipping the LLM core with a structured control loop, thereby enabling dynamic reasoning, adaptive long-horizon planning, multi-hop information retrieval, iterative tool use, and the generation of structured analytical reports \cite{huang2025deepresearchagentssystematic, wang2023survey, yao2023react}.

The introduction of agency relies on three primary functional pillars: (1) Planning and Control, typically achieved through mechanisms like iterative planning and self-reflection to maintain task alignment and correct errors \cite{park2023generative}; (2) External Tool Utilization, encompassing code interpreters, specialized APIs, and web search to access computational and real-time resources \cite{schick2023toolformer}; and (3) Contextual Grounding, most commonly implemented via Retrieval-Augmented Generation (RAG) \cite{lewis2020retrieval} to seamlessly incorporate up-to-date or proprietary domain-specific data. This composite framework allows agents to interact dynamically with complex environments and accumulate long-term memory, thereby achieving sustained logical reasoning and verifiable evidence retrieval necessary for accelerated research, literature review synthesis, and scientific hypothesis generation.

However, evaluating the true scientific utility of these systems remains a significant challenge. Crucially, the effective assessment of these domain-specific agents, particularly in high-stakes fields like Biology and Medicine, necessitates challenging benchmarks that test deep comprehension, multi-step causal reasoning, and faithful evidence extraction over highly specialized and voluminous literature \cite{biomedical_benchmark}, moving beyond general knowledge assessment.

\subsection{Scientific Benchmarks}
General scientific benchmarks have emerged to evaluate AI systems across disciplines like physics, chemistry, and biology\cite{zhao2025superchemmultimodalreasoningbenchmark, qiu2025phybenchholisticevaluationphysical, peng2025proof2hybridautomaticmathematicalbenchmark}, with a focus on testing domain knowledge and problem-solving skills. Initial efforts focused on high-difficulty, factual question-answering, exemplified by \textbf{GPQA} \cite{Rein2023GPQAAG} and its scaled successor \textbf{SuperGPQA} \cite{MAPTeam2025SuperGPQA}, which feature expert-authored, graduate-level questions. Other highly challenging benchmarks include \textbf{HLE} \cite{Phan2025HLE}, which assesses advanced STEM and humanities through short-answer and multimodal tasks, and \textbf{R-Bench} \cite{wu2024rbench}, which targets Graduate/Olympiad-level reasoning using bilingual, multimodal inputs. Furthermore, \textbf{SciEval} \cite{scibench_2023} aggregates academic datasets for broad STEM coverage, while \textbf{OlympicArena} \cite{Huang2024OlympicArena} evaluates multidisciplinary cognitive reasoning with a focus on process-level assessment.

\subsection{Biology-Specific Benchmarks}
Biology-specific benchmarks have focused on subfield-specific tasks, but few address the integrated experimental reasoning that is critical for advancing the field. Table \ref{tab:bio-benchmarks-comparison} summarizes key biology benchmarks and their limitations relative to \textbf{BABE}. Several benchmarks are Sequence-Centric. For example, benchmarks like Biology-Instructions \cite{he2025biologyinstructionsdatasetbenchmarkmultiomics} and ProteinBench \cite{ye2024proteinbenchholisticevaluationprotein} focus on sequence-based tasks (e.g., DNA sequence alignment, protein secondary structure prediction). Understanding structures of Bio-macromolecules is a crucial task in biology. Hence, there are Structure-Centric Benchmarks, especially in protein area. For example, ProteinShake\cite{Kucera2023ProteinShake} and PepPCBench \cite{Zhai2025PepPCBench} evaluate models on protein structure, using PDB-formatted structures to assign proteins to fold families. While these benchmarks are valuable for assessing structural biology models \cite{alphafold3_2024}), they do not require models to interpret experimental data related to structures. Meanwhile, there is a small number of biology benchmarks incorporate multiple modalities, but they lack the depth of experimental reasoning required for real research. For example, BioASQ \cite{bioasq_2015} organizes challenges on biomedical semantic indexing and question answering relevant to hierarchical text classification, machine learning, information retrieval, QA from texts and structured data, multi-document summarization and many other areas. OlymBench is a new benchmark designed to evaluate the reasoning ability of large language models, with challenges reaching competition difficulty. 
All existing biology-specific benchmarks fail to address one or more critical needs: 
\begin{itemize}
    \item Real experimental data: Most use simplified data or summarized data rather than figures and datasets from published papers.
    \item Integrated reasoning: Tasks do not require linking experimental results to contextual background;
    \item Broad domain coverage: Benchmarks are limited to single subfields rather than spanning multiple biological domains.
\end{itemize}
BABE addresses these gaps by focusing on research-derived, multimodal tasks that require the same reasoning as practicing biologists.
\begin{table*}[t]
    \centering
    \caption{Key Biology Benchmarks and their Gaps Relative to \textbf{BABE}}
    \label{tab:bio-benchmarks-comparison}
    \adjustbox{max width=\linewidth}{
        \begin{tabular}{l|c|c|c|l}
            \toprule
            \textbf{Benchmark} & \textbf{Real Experimental Data} & \textbf{Integrated Reasoning} & \textbf{Domain Coverage} & \textbf{Primary Focus} \\
            \midrule
            ProteinBench & No & Limited & Narrow & Computational Metrics \\
            ProteinShake & No & Limited & Narrow & Standardized Structural Data \\
            PepPCBench & No & Limited & Narrow & Structure Prediction Accuracy \\
            BioASQ & No & Medium & Broad & Text/Sequence QA Corpus \\
            OlymBench & No & High & Broad & High-difficulty Logical Deduction \\ 
            \midrule
            \textbf{BABE} & \textbf{Yes} & \textbf{High} & \textbf{Broad} & \textbf{Research-Derived Multimodal Tasks} \\
            \bottomrule
        \end{tabular}
    }
\end{table*}
\section{Approach}

\subsection{Problem Formulation}

The \textbf{Biology Arena Benchmark (BABE)} is introduced as a novel diagnostic framework designed to rigorously evaluate Large Language Models (LLMs) on complex reasoning tasks within the biomedical domain, specifically over a single source research document $D$. Unlike existing benchmarks, which often emphasize factual recall or single-hop retrieval, fail to fully diagnose model robustness in \textit{compositional reasoning}, our benchmark focus on handling the complexity and quantitative content inherent in scientific literature. BABE addresses this limitation by defining a structured question triplet:
\[
Q_{\text{BABE}} = \{ Q_1, Q_2, Q_3 \},
\]
where the logical relationship between consecutive questions, $R(Q_i, Q_{i+1})$, is formally classified into two distinct diagnostic categories: \textbf{Strong Correlation} ($R_{\text{Strong}}$) and \textbf{Weak Correlation} ($R_{\text{Weak}}$). This structured methodology is essential for precisely measuring LLM \textit{depth} (sequential inference) and \textit{breadth} (parallel extraction) of understanding in a domain-specific context.

\bigskip
The problem formulation is centered on the interdependency within the question set $Q_{\text{BABE}}$. The overall type of a benchmark instance is defined by the combined logical relations:
\[
Q_{\text{BABE}}^{\text{Type}} \equiv R(Q_1, Q_2) \wedge R(Q_2, Q_3).
\]
The \textbf{Strong Correlation} relation ($R_{\text{Strong}}$) captures sequential, multi-hop reasoning, where the derived output of a preceding question is a necessary input for the subsequent one. This relationship is formally defined as:
\[
R_{\text{Strong}} \;\Leftrightarrow\; \forall i,j (i<j), \; A_j \text{ requires } A_{i} \text{ for optimal derivation from } D.
\]
Conversely, the \textbf{Weak Correlation} relation ($R_{\text{Weak}}$) captures parallel, independent extraction, where questions are logically uncoupled, testing the model's ability to maintain multiple distinct contexts simultaneously. This is formalized as:
\[
R_{\text{Weak}} \;\Leftrightarrow\; \exists C_i \subset D \; \text{s.t.} \; A_i = \text{Extract}(C_i), \; \text{and} \; \forall i \neq j, \; C_i \cap C_j \approx \emptyset.
\]
The formal constraint of $R_{\text{Strong}}$ diagnoses failure modes related to error propagation and reasoning drift in \textit{Chain-of-Thought (CoT)} processes, while $R_{\text{Weak}}$ diagnoses issues such as semantic interference during simultaneous knowledge retrieval.

\subsection{Data Collection}

To construct a high-quality benchmark tailored for rigorous evaluation of domain-specific reasoning, we adopted a multi-stage data collection pipeline integrating frontier literature curation, expert-driven item development, and structured quality assurance. 

As shown in \ref{fig:main-fig}(A), We first curated a corpus of cutting-edge scientific materials, including recently published peer-reviewed papers, domain-specific monographs, and authoritative review articles. The selection criteria emphasized (i) recency of publication, (ii) relevance to the target scientific domain, and (iii) conceptual depth suitable for assessing multi-step reasoning. Only sources meeting all criteria were retained for downstream question construction.

For each selected paper or book chapter, domain experts generated a set of three assessment items. The items were designed to probe different cognitive dimensions, including conceptual understanding, methodological interpretation, and higher-order reasoning. All questions were required to be self-contained, unambiguous, and faithful to the source material while avoiding superficial fact-retrieval prompts.

A secondary panel of senior experts conducted a rigorous review of all drafted items. The review served two purposes:

\begin{itemize}
    \item \textbf{Relevance Assessment}: Each question was labeled as either \emph{strong correlation} or \emph{weak correlation} to the core knowledge unit extracted from the source text. Strongly related questions directly test key concepts and reasoning chains presented in the material, whereas weakly related questions assess peripheral or contextual understanding.
    \item \textbf{Correctness Verification}: Reviewers evaluated the factual fidelity, logical coherence, and answer correctness for every item.
\end{itemize}

Items passing both relevance assessment and correctness verification were accepted into the benchmark. Items found to contain factual inaccuracies, ambiguous phrasing, or misalignment with the source were returned to the original authors for revision. After revision, questions underwent a second-round review before being considered for final inclusion. With the help of LLMs, simple questions were removed by the reviewers in this round.

Only items that successfully completed the full development--review--revision cycle were incorporated into the released benchmark. This multi-layered pipeline ensures that the dataset captures high-quality, expert-vetted questions with explicit relevance annotations, enabling more fine-grained evaluation of model capabilities across varying levels of semantic alignment with the underlying scientific sources.

\section{Experiments}

\subsection{Overall Performance Analysis}
\begin{table}[htbp]
    \centering
    \caption{Model Performance on BABE Comparison}
    \adjustbox{max width=0.95\linewidth}{
    \begin{tabular}{l|c|c|c}
    \hline
    \textbf{Model Name} & \textbf{Average Score} & \textbf{Strong Correlation} & \textbf{Weak Correlation} \\
    \hline
    OpenAI-GPT-5.1-high & \textbf{52.31} & \textbf{51.79} & 52.86 \\
    Gemini-3-Pro-Preview-Exp & 52.02 & 49.05 & \textbf{55.16} \\
    OpenAI-o3-high.code & 51.62 & 51.22 & 52.05 \\
    Doubao-1.8-1228 & 43.27 & 43.38 & 43.16 \\ 
    Gemini-2.5-Pro & 42.17 & 42.02 & 42.33 \\
    Claude-Sonnet-4.5-thinking-azure & 41.93 & 41.94 & 41.91 \\
    Doubao-1.6-1015-high.foreval & 39.84 & 40.70 & 38.94 \\
    Doubao-1.6-thinking.0715.foreval & 39.54 & 39.67 & 39.40 \\
    Doubao-1.6-1015-high.foreval (A) & 39.34 & 35.10 & 43.81 \\
    Claude-Opus-4.1-thinking-azure & 37.35 & 36.30 & 38.46 \\
    OpenAI-gpt4.1-0414 & 36.86 & 32.61 & 41.34 \\
    Doubao-1.6-1015-high.foreval (B) & 36.64 & 34.88 & 38.49 \\
    QwenAPI-vl-max.latest & 32.81 & 32.89 & 32.72 \\
    Doubao-1.5-pro-thinking.vision.0428.eval & 31.71 & 32.12 & 31.28 \\
    Claude-Sonnet-4.5-nothinking-azure & 31.66 & 33.67 & 29.55 \\
    QwenAPI-3-max-0923 & 26.54 & 27.70 & 25.32 \\
    GPT4o-1120 & 23.93 & 20.80 & 27.22 \\
    Doubao-1.6-flash.0828.foreval & 22.04 & 20.34 & 23.83 \\
    GLM-4.5-V & 20.83 & 18.09 & 23.72 \\
    \hline
    \end{tabular}
    }
    \label{tab:model_performance}
\end{table}

Table~\ref{tab:model_performance} presents a comprehensive comparison of different models on the BABE benchmark under both strong and weak correlation settings.
Overall, \textbf{OpenAI-GPT-5.1-high} achieves the best performance, obtaining the highest average score of \textbf{52.31}, and demonstrating consistently strong results across both strong (51.79) and weak (52.86) correlation subsets. This indicates robust reasoning capabilities that generalize well across varying dependency structures.

Models in the lower performance tier generally struggle with both correlation settings, with notable variations. For example, \textbf{OpenAI-gpt4.1-0414} (36.86) performs better in weak correlation (41.34) than strong correlation (32.61), while \textbf{Claude-Sonnet-4.5-nothinking-azure} (31.66) shows the opposite trend (33.67 for strong correlation vs. 29.55 for weak correlation), indicating divergent design trade-offs in handling explicit vs. implicit reasoning. The lowest-performing models, such as \textbf{GLM-4.5-V} (20.83) and \textbf{Doubao-1.6-flash.0828.foreval} (22.04), score consistently low across both subsets, suggesting fundamental limitations in reasoning capabilities for the BABE benchmark’s task requirements.

\subsection{Strong vs. Weak Correlation}
We observe notable performance differences between strong and weak correlation scenarios for several models.
For example, \textbf{Gemini-3-Pro-Preview-Exp} exhibits a clear advantage under weak correlation conditions (55.16), substantially outperforming its strong correlation score (49.05). This suggests that the model is particularly effective when explicit logical dependencies are reduced, potentially benefiting from broader contextual reasoning.
In contrast, models such as \textbf{Claude-Sonnet-4.5-thinking-azure} and \textbf{Gemini-2.5-Pro} show highly balanced performance across the two settings, indicating stable behavior regardless of correlation strength.


\subsection{Reasoning Behavior Analysis on BABE}
\begin{figure*}[htbp]
    \centering
    \includegraphics[width=0.98\linewidth]{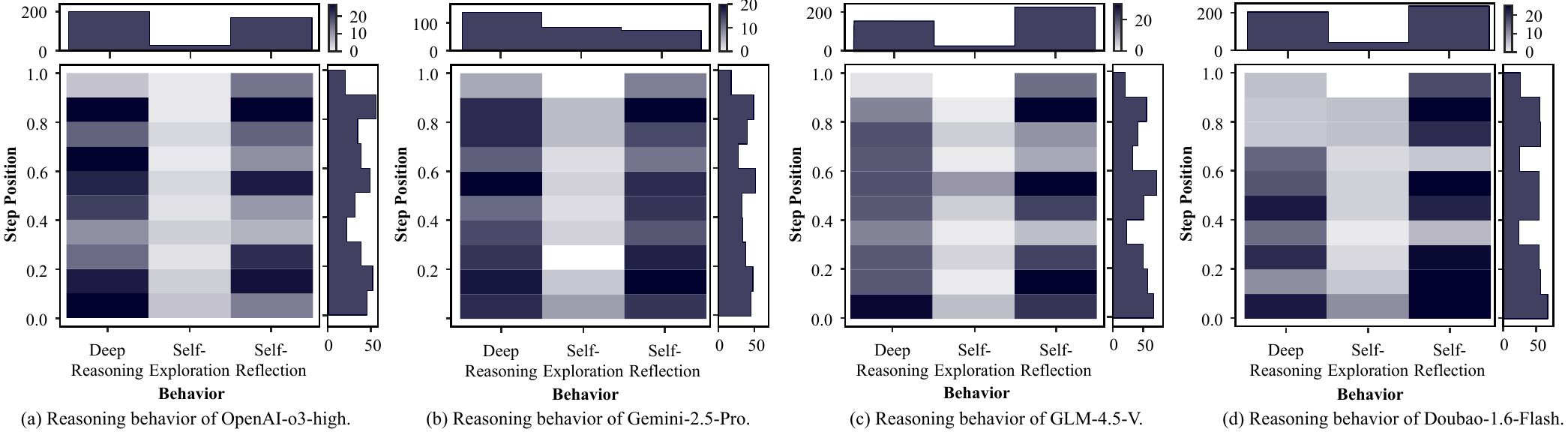}
    \caption{The Reasoning Behavior Distribution on BABE across four LLMs.}
    \label{fig:reasoning-behavior}
\end{figure*}

\paragraph{\textbf{BABE requires deeper reasoning.}}
To better understand the reasoning demands of BABE, we compare models at the two extremes of performance: the two best-performing models and the two worst-performing models. Figure~\ref{fig:reasoning-behavior} shows a clear association between success on BABE and the prevalence of {Deep Reasoning} behaviors during inference. Specifically, higher-performing models devote a substantially larger portion of their inference steps to deep reasoning, whereas lower-performing models exhibit a much smaller share of such behaviors. This pattern indicates that BABE is not primarily solved by shallow pattern matching; instead, it rewards deeper reasoning that resolves implicit or non-trivial dependencies in the input.

\paragraph{\textbf{Excessive self-reflection on BABE can lead to a substantial degradation in reasoning performance.}}
Figure~\ref{fig:reasoning-behavior} also reveals a second, more subtle failure mode among the worst-performing models: they exhibit {episodic Self-Reflection} at a notably higher rate, often exceeding their own proportion of deep reasoning. While limited self-reflection can be beneficial, these results suggest that frequent, repeated self-reflection without commensurate progress in deep reasoning is associated with worse outcomes on BABE. A plausible interpretation is that weaker models fall into an ``overthinking'' loop—spending many steps reconsidering intermediate thoughts or reformulating the approach—yet failing to advance the core reasoning needed to reach correct conclusions. In practice, this behavior consumes the inference budget and increases the chance of drifting away from relevant evidence on BABE, which ultimately harms accuracy.

\paragraph{\textbf{Strong performance on BABE depends on sustained, evenly applied deep reasoning.}}
Beyond overall proportions, we observe that strong models tend to maintain deep reasoning consistently throughout the inference trajectory. In contrast, some models may begin with deep reasoning but gradually reduce such behaviors later in the process, yielding a sparser and less stable reasoning pattern. Figure~\ref{fig:reasoning-behavior} suggests that this ``early burst'' is insufficient for BABE: correctly solving BABE examples often requires repeatedly revisiting earlier premises, integrating newly derived implications, and maintaining coherent multi-step constraints until a final decision is justified. Therefore, strong performance on BABE depends not only on initiating deep reasoning, but on sustaining it in a relatively uniform manner across the full inference process.

\subsection{Convergence Score Analysis}

\begin{figure}[htbp]
    \centering
    \includegraphics[width=0.9\linewidth]{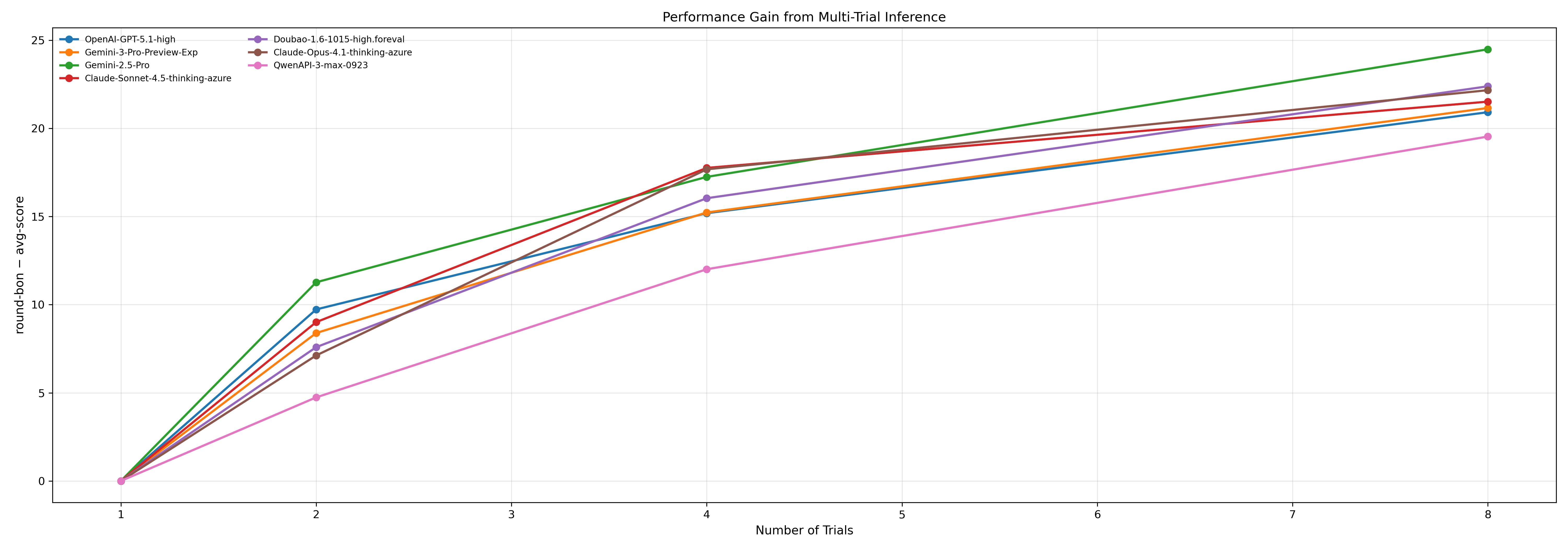}
    \caption{
    Performance gain from multi-trial inference.
    }
    \label{fig:gain_curve}
\end{figure}

We define the gain of multi-trial inference as:
\[
\text{Gain}(n) = \text{model-BoN}(n) - \text{avg-score}(n),
\]
where $BoN$ is short for "Best of N",  $n$ denotes the number of inference trials.
This metric isolates the improvement obtained purely from repeated sampling and aggregation, independent of the model’s average single-run performance.

To estimate the saturation behavior of gain, we fit a parametric saturating function:
\[
\text{Gain}(n) = a \cdot \left(1 - e^{-b n}\right),
\]
where $a$ represents the asymptotic (converged) gain and $b$ controls the convergence speed.
Parameters are estimated using nonlinear least squares on observed gains at $n=\{1,2,4,8\}$.

Figure~\ref{fig:gain_curve} shows the gain achieved by increasing the number of inference trials.
All models exhibit a monotonic increase in gain, confirming that multi-trial inference consistently improves performance beyond expected single-run outcomes.
However, the slope of the curves decreases as $n$ grows, indicating diminishing marginal returns and the onset of convergence.

\begin{table*}[htbp]
\centering
\small
\setlength{\tabcolsep}{3pt}
\adjustbox{max width=\textwidth}{
\begin{tabular}{c|cc|cc|cc|cc|cc|cc}
\toprule
\textbf{Trial} 
& \multicolumn{2}{c|}{\textbf{OpenAI-GPT-5.1}}
& \multicolumn{2}{c|}{\textbf{Gemini-3}}
& \multicolumn{2}{c|}{\textbf{Gemini-2.5-pro}}
& \multicolumn{2}{c|}{\textbf{Claude-S-4.5}}
& \multicolumn{2}{c|}{\textbf{Doubao-1.6}}
& \multicolumn{2}{c|}{\textbf{Claude-O-4.1}}
\\
\cmidrule(lr){2-13}
& Avg & BoN
& Avg & BoN
& Avg & BoN
& Avg & BoN
& Avg & BoN
& Avg & BoN
 \\
\midrule
1
& 52.31 & 52.31
& 52.02 & 52.02
& 42.17 & 42.17
& 41.93 & 41.93
& 39.84 & 39.84
& 37.35 & 37.35
 \\

2
& 54.14 & 63.87
& 54.16 & 62.55
& 45.06 & 56.33
& 43.18 & 52.19
& 36.70 & 46.29
& 37.03 & 44.15
 \\

4
& 53.87 & 69.06
& 54.32 & 69.55
& 45.14 & 62.39
& 42.23 & 59.99
& 37.00 & 53.04
& 37.31 & 54.99
 \\

8
& 55.09 & 76.01
& 53.59 & 74.75
& 45.14 & 69.63
& 42.60 & 64.12
& 37.58 & 59.97
& 36.35 & 58.52
 \\

\midrule
\textbf{Convergence}
& -- & 81.90
& -- & 86.54
& -- & 81.31
& -- & 71.78
& -- & 80.77
& -- & 73.73
 \\
\bottomrule
\end{tabular}
}
\caption{
Raw performance under multi-trial inference.
Rows correspond to different trial numbers and the predicted asymptotic convergence value.
For each model, we report the average score (Avg) and round-based score (Round).
The convergence value is estimated by fitting a saturating exponential model
using trial results at $n \in \{1,2,4,8\}$.
}
\label{tab:trial-convergence-results}
\end{table*}

Strong reasoning models such as \textbf{OpenAI-GPT-5.1-high} and \textbf{Gemini-3-Pro-Preview-Exp} exhibit relatively fast convergence, with predicted asymptotic gains around 30 points.
These models show limited additional improvement beyond a moderate number of trials, suggesting that their reasoning quality is already robust in single or few-shot inference.

In contrast, several mid-tier models (e.g., \textbf{Gemini-2.5-Pro} and \textbf{Claude-Opus-4.1-thinking-azure}) display higher estimated gain limits, exceeding 35 points.
This indicates greater diversity in their generated reasoning trajectories, allowing aggregation to recover substantially better solutions despite weaker average performance.

Taken together, these results indicate that success on BABE generally requires at least 4–6 inference trials even for frontier models, and 8+ trials for most non-frontier models, highlighting the intrinsic difficulty of experimental reasoning tasks and the limitations of single-pass inference.

\section{Conclusion}

In this paper, we introduced BABE (Biology Arena BEnchmark), a novel and rigorous benchmark designed to evaluate the experimental reasoning capabilities of large language models in biology. Unlike existing benchmarks that emphasize factual recall, sequence-level prediction, or isolated subtasks, BABE focuses on a core competency of real biological research: integrating experimental results with contextual background to derive scientifically meaningful conclusions.

BABE is constructed entirely from peer-reviewed research papers and real-world biological studies, ensuring that its tasks faithfully reflect the complexity, interdisciplinarity, and ambiguity inherent in actual scientific inquiry. By organizing questions into structured triplets with explicitly defined strong and weak correlation relationships, BABE provides a fine-grained diagnostic framework for assessing both sequential multi-hop reasoning and parallel information extraction within a single source document. This formulation enables more precise identification of reasoning failure modes that are often obscured in conventional benchmarks.

Overall, BABE fills a critical gap in the evaluation of biological AI systems by moving beyond task-level competence toward research-level reasoning. We believe BABE will serve as a valuable testbed for future advances in biologically grounded LLMs, deep research agents, and multimodal scientific reasoning systems. Looking forward, we hope this benchmark will encourage the development of models that reason more like practicing scientists—grounding their conclusions in evidence, maintaining coherence across experimental contexts, and ultimately contributing more reliably to real biological discovery.

\clearpage

\bibliographystyle{plainnat}
\bibliography{main}

\clearpage

\beginappendix

\section{Example Questions of BABE}

\begin{figure}[h!]
    \centering 

    \includegraphics[width=0.80\textwidth]{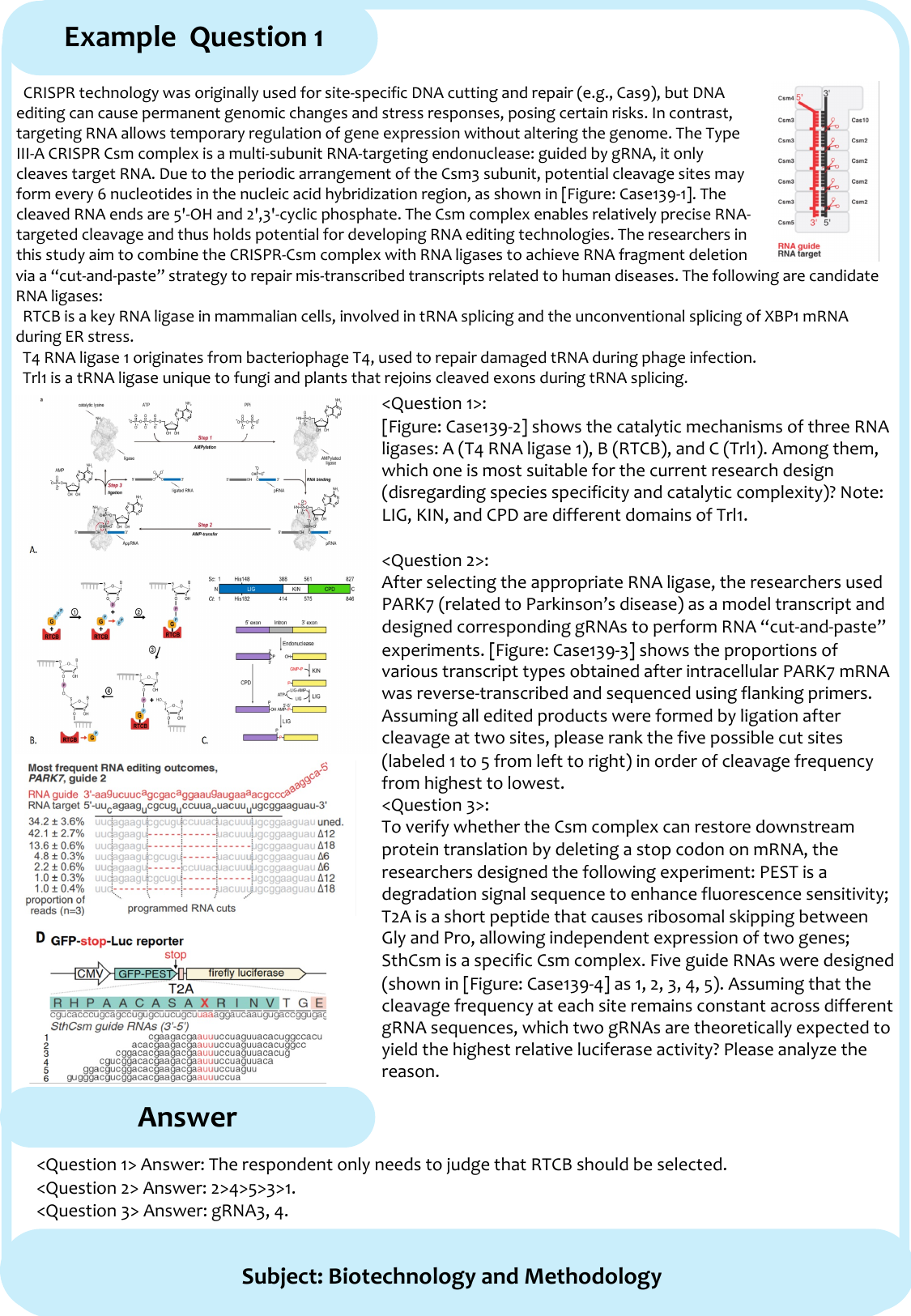} 
    \caption{Example Question 1 of BABE} 
\end{figure}
\begin{figure}[h!]
    \centering 

    \includegraphics[width=0.80\textwidth]{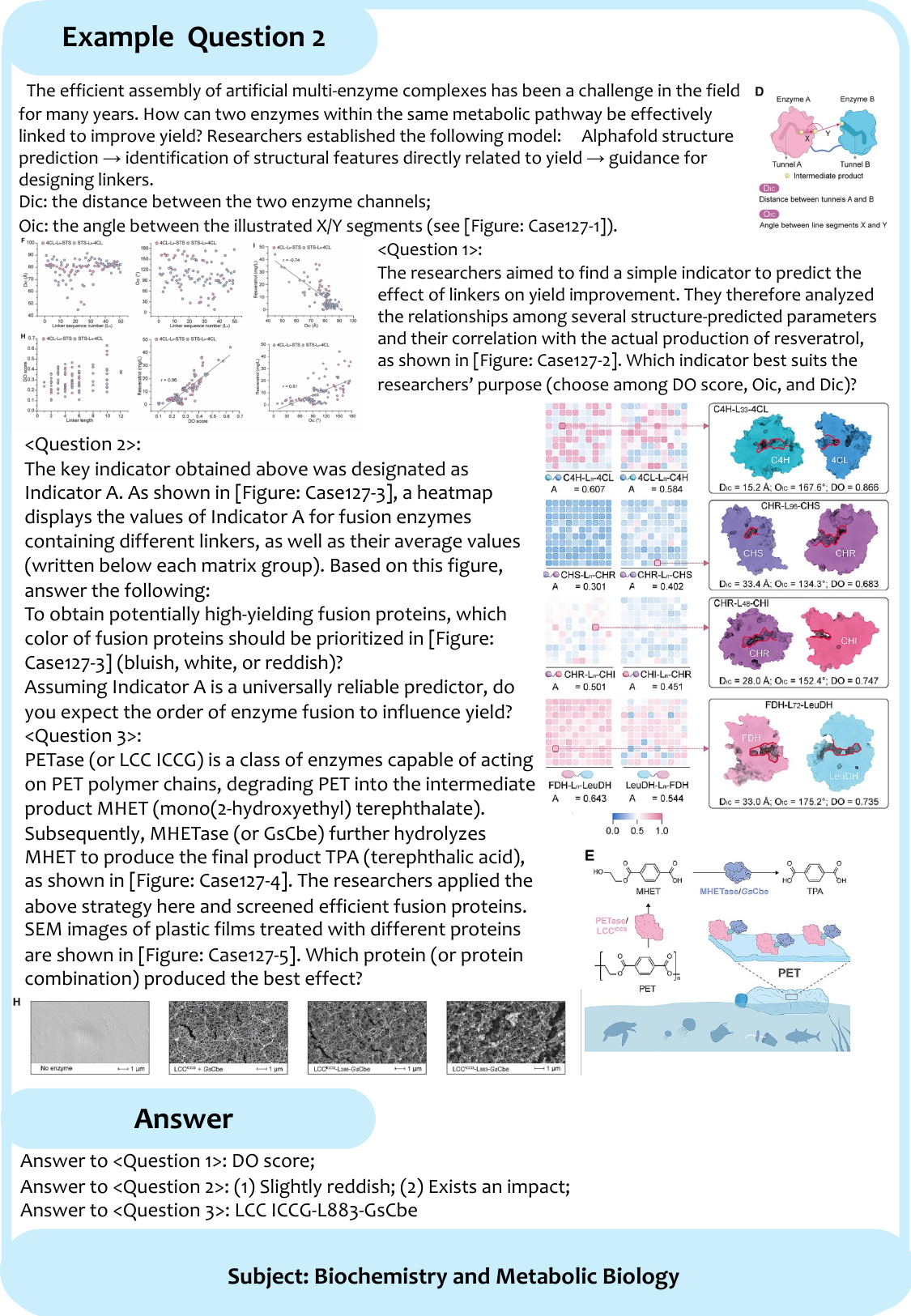} 
    \caption{Example Question 2 of BABE} 
\end{figure}
\begin{figure}[h!]
    \centering 

    \includegraphics[width=0.80\textwidth]{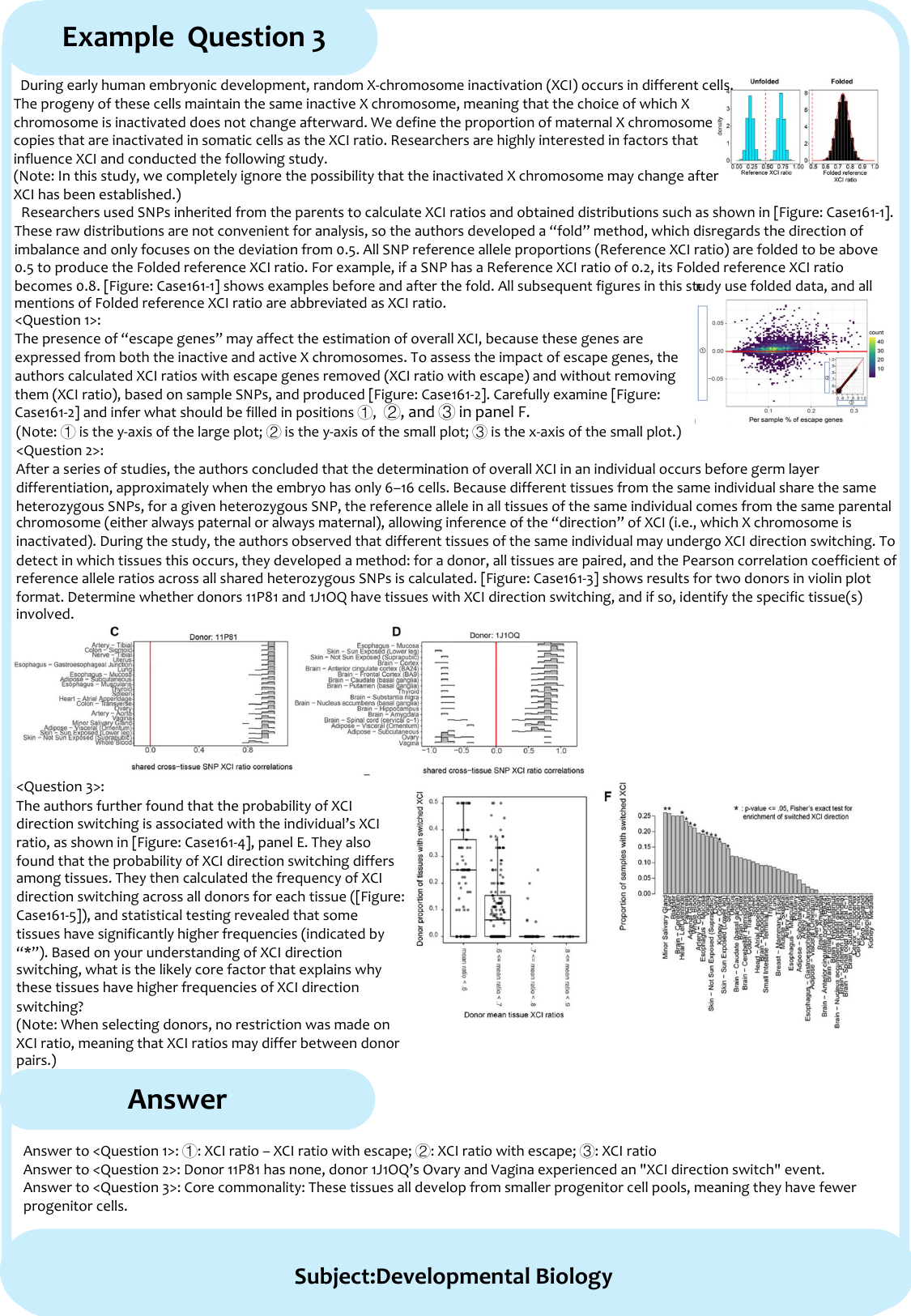} 
    \caption{Example Question 3 of BABE} 
\end{figure}
\begin{figure}[h!]
    \centering 

    \includegraphics[width=0.80\textwidth]{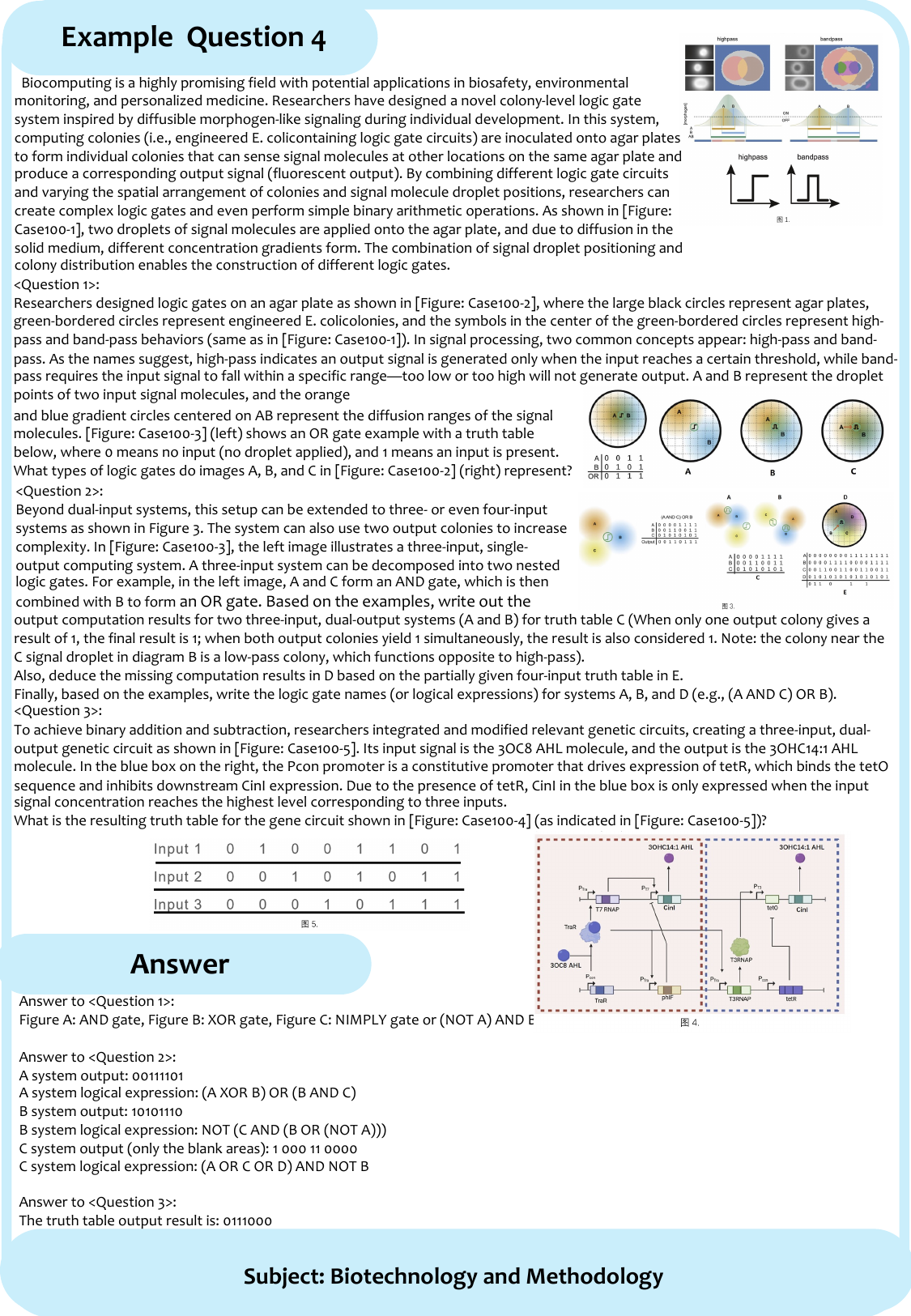} 
    \caption{Example Question 4 of BABE} 
\end{figure}
\begin{figure}[h!]
    \centering 

    \includegraphics[width=0.80\textwidth]{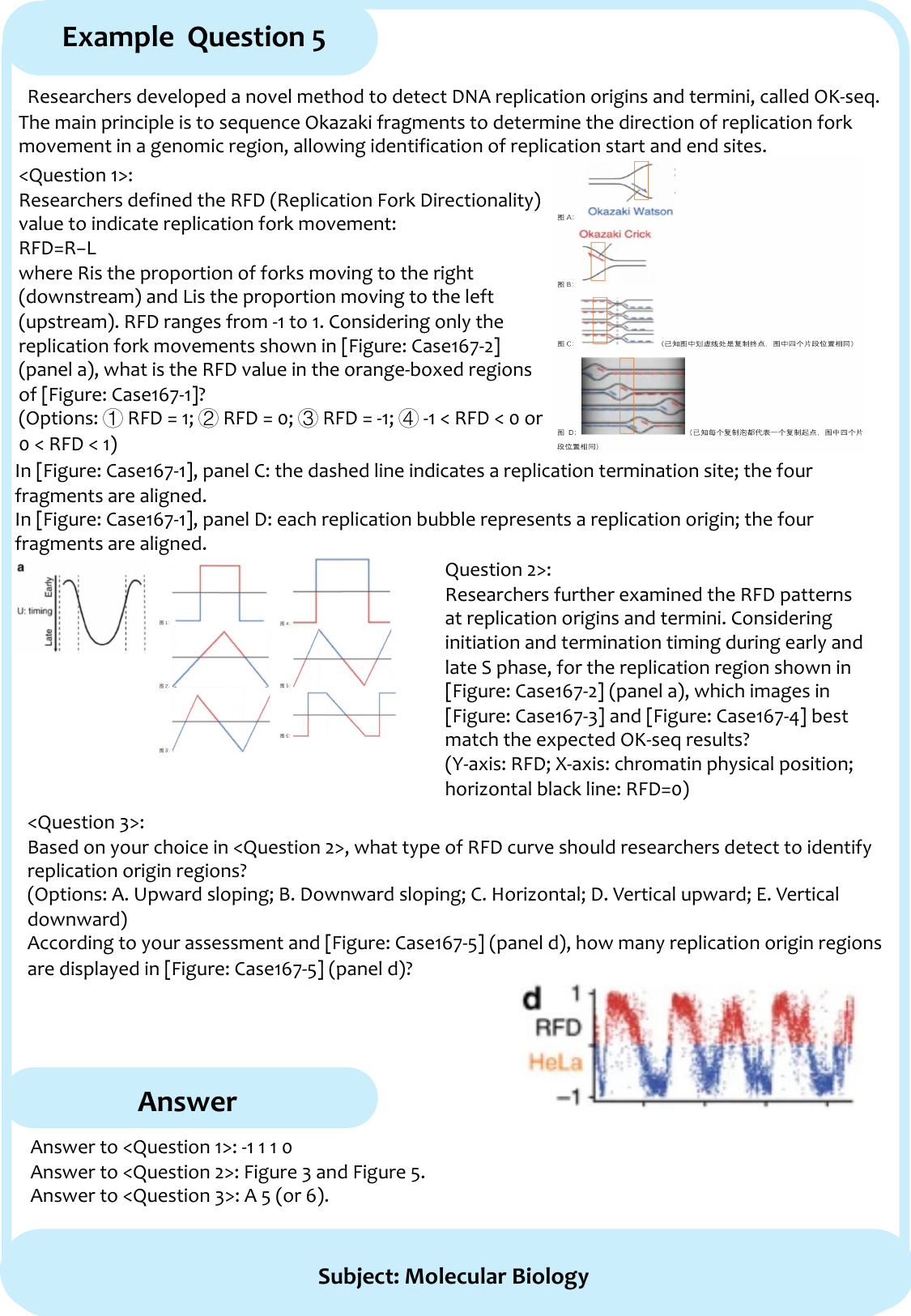} 
    \caption{Example Question 5 of BABE} 
\end{figure}
\begin{figure}[h!]
    \centering 

    \includegraphics[width=0.80\textwidth]{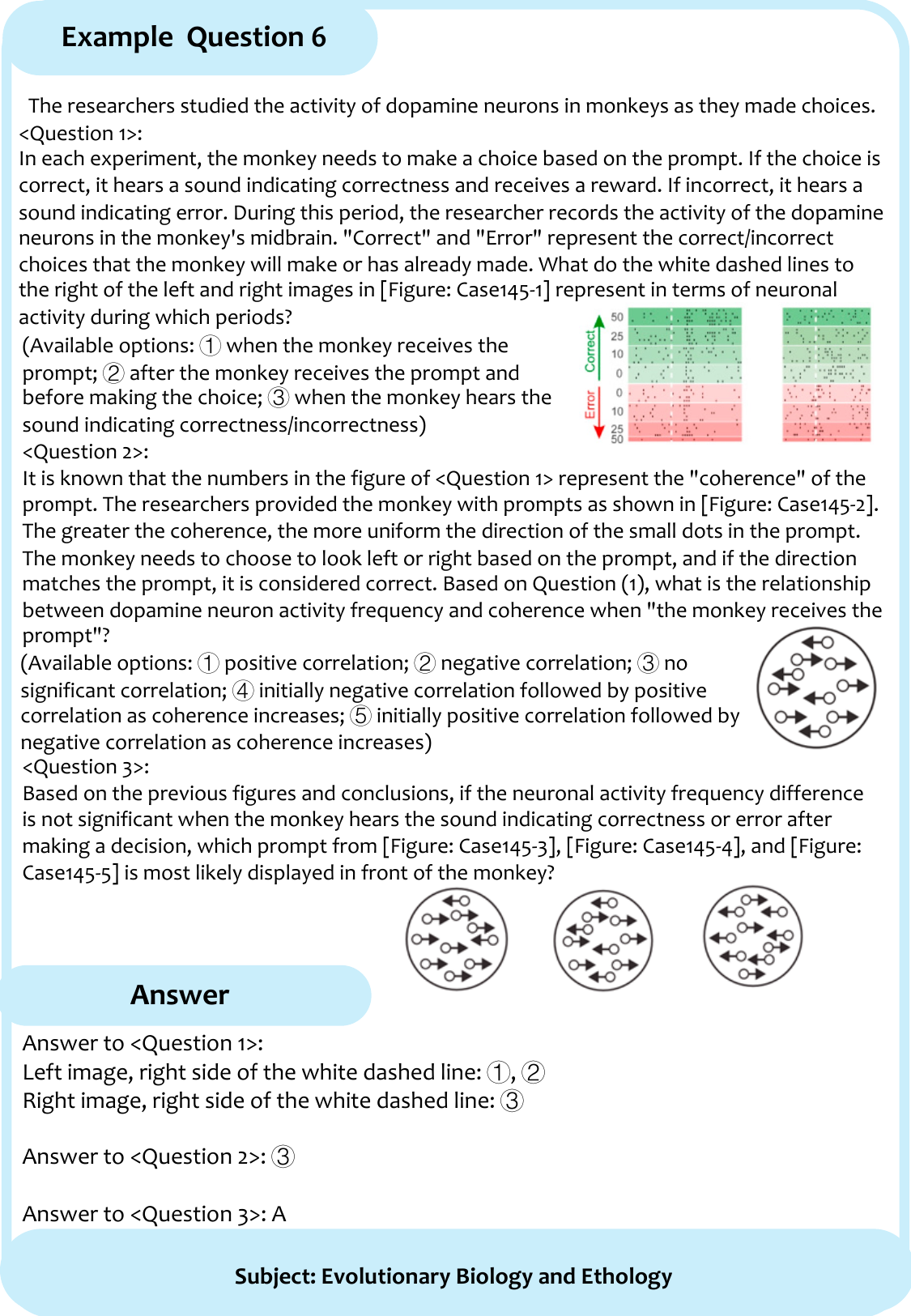} 
    \caption{Example Question 6 of BABE} 
\end{figure}
\begin{figure}[h!]
    \centering 

    \includegraphics[width=0.80\textwidth]{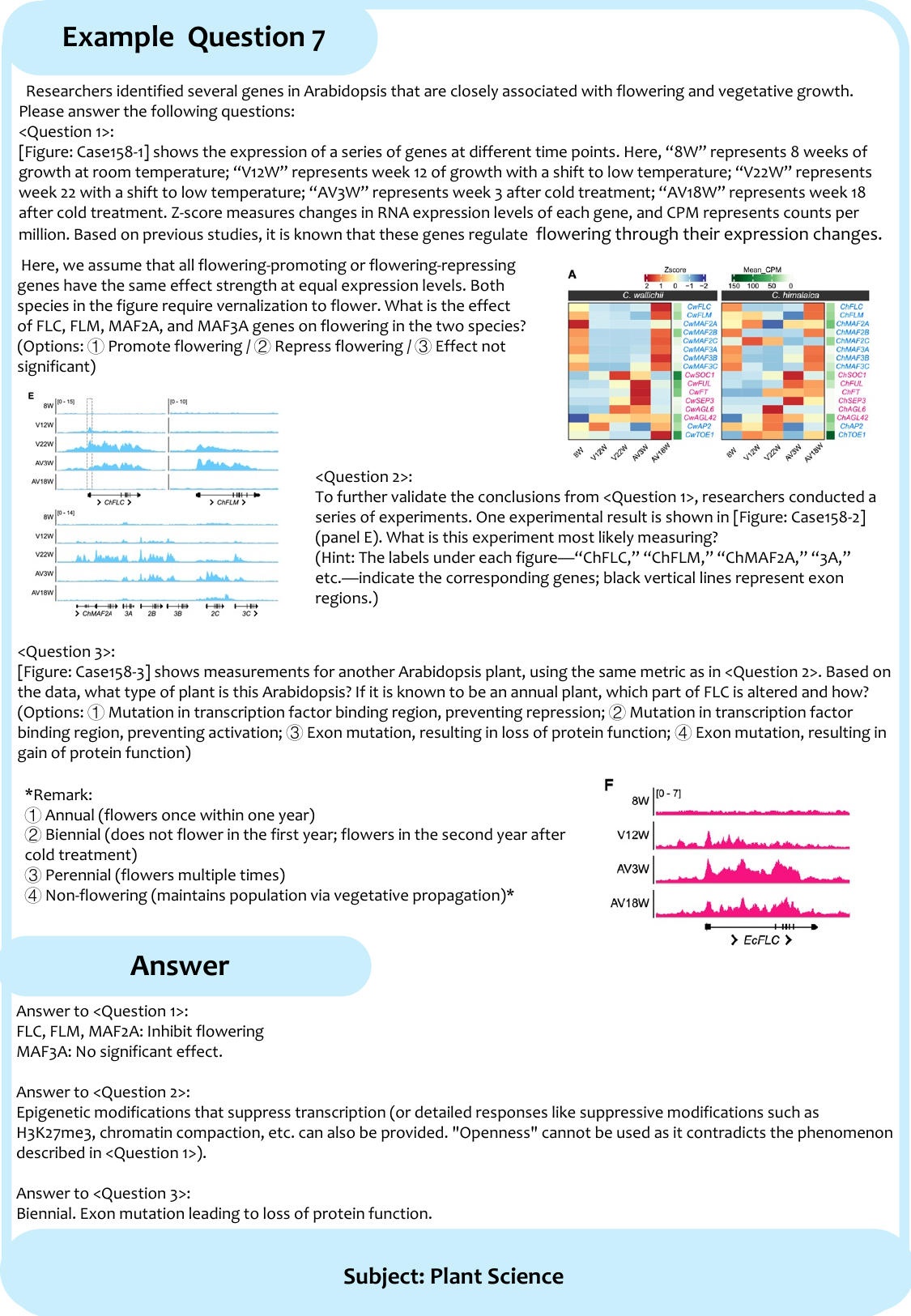} 
    \caption{Example Question 7 of BABE} 
\end{figure}
\begin{figure}[h!]
    \centering 

    \includegraphics[width=0.80\textwidth]{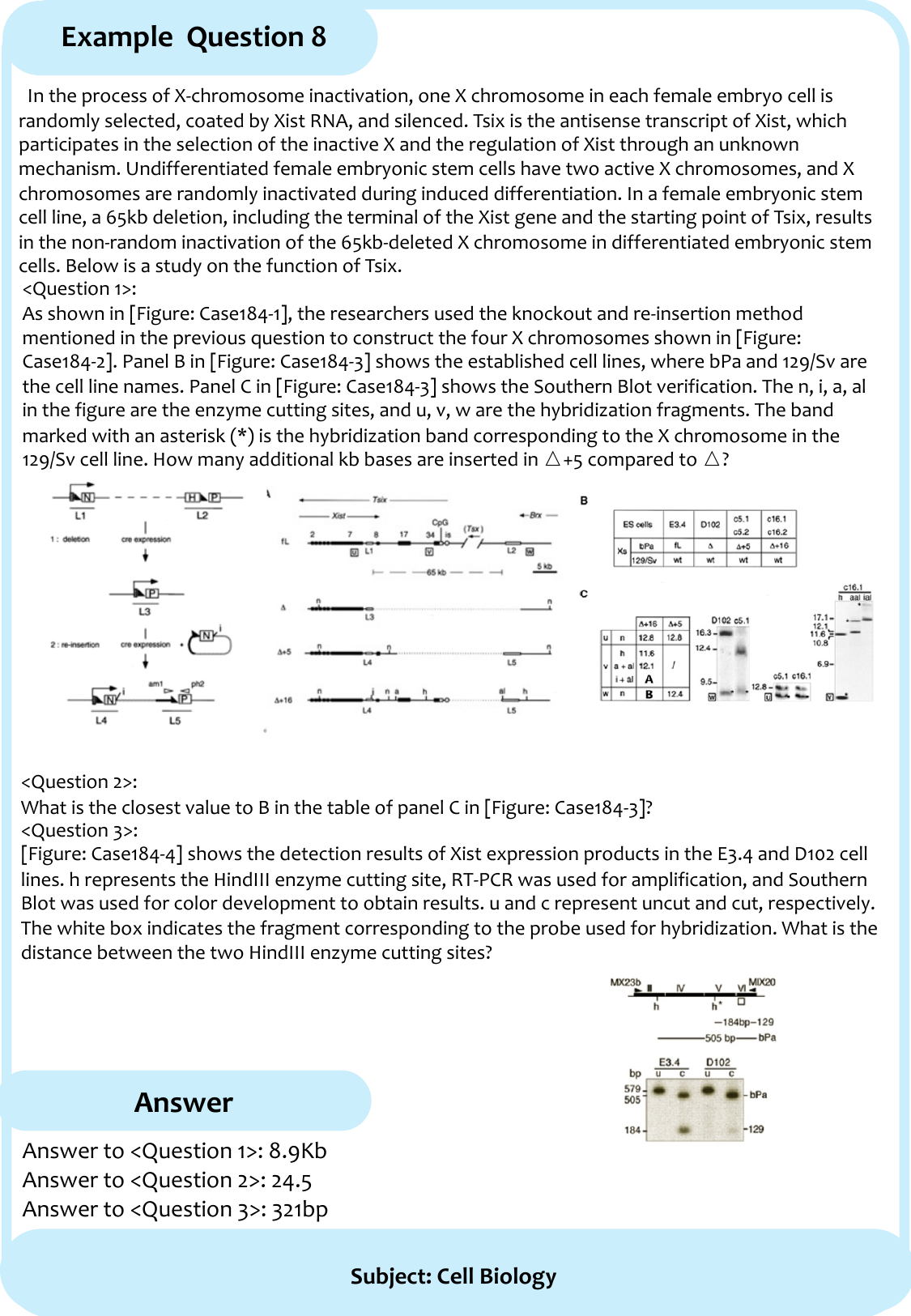} 
    \caption{Example Question 8 of BABE} 
\end{figure}

\end{document}